\documentclass[letterpaper, 10 pt, conference]{ieeeconf}  
\usepackage{moreverb,url}
\usepackage{bm}
\usepackage[colorlinks,bookmarksopen,bookmarksnumbered,citecolor=red,urlcolor=red]{hyperref}
\usepackage{dsfont}
\IEEEoverridecommandlockouts                              
\usepackage{graphicx} 
\usepackage{amsmath} 
\usepackage{subfig}
\title{\LARGE \bf
A Soft e-Textile Sensor for Enhanced Deep Learning-based\\ Shape Sensing of Soft Continuum Robots
}

\author{Eric Vincent Galeta$^{*1}$, Ayman A. Nada$^{1}$, Sabah M. Ahmed$^{2}$, Victor Parque$^{3}$, Haitham El-Hussieny$^{*1}$
\thanks{*This work was not supported by any organization}
\thanks{$^{1}$Department of Mechatronics and Robotics, Egypt-Japan University of Science and Technology, Alexandria, Egypt
        {\tt\small haitham.elhussieny@ejust.edu.eg}}%
        \thanks{$^{2}$Electrical Engineering Department,  Faculty of Engineering, Assiut University, Egypt}%
\thanks{$^{3}$Department of Modern Mechanical Engineering, Waseda University, Tokyo, Japan}%
}
\graphicspath{{Pictures/}}

\begin{document}

\maketitle
\thispagestyle{empty}
\pagestyle{empty}

\begin{abstract}
The safety and accuracy of robotic navigation hold paramount importance, especially in the realm of soft continuum robotics, where the limitations of traditional rigid sensors become evident. Encoders, piezoresistive, and potentiometer sensors often fail to integrate well with the flexible nature of these robots, adding unwanted bulk and rigidity. To overcome these hurdles, our study presents a new approach to shape sensing in soft continuum robots through the use of soft e-textile resistive sensors. This sensor, designed to flawlessly integrate with the robot's structure, utilizes a resistive material that adjusts its resistance in response to the robot's movements and deformations. This adjustment facilitates the capture of multidimensional force measurements across the soft sensor layers. A deep Convolutional Neural Network (CNN) is employed to decode the sensor signals, enabling precise estimation of the robot's shape configuration based on the detailed data from the e-textile sensor. Our research investigates the efficacy of this e-textile sensor in determining the curvature parameters of soft continuum robots. The findings are encouraging, showing that the soft e-textile sensor not only matches but potentially exceeds the capabilities of traditional rigid sensors in terms of shape sensing and estimation. This advancement significantly boosts the safety and efficiency of robotic navigation systems.
\end{abstract}

\section{INTRODUCTION}

The advent of soft robotics signifies a profound shift in the robotics paradigm, providing unmatched flexibility and safety over traditional rigid robots \cite{el2020nonlinear, seleem2023recent}. In contrast to conventional robots, which are constrained by their rigid structures, soft robots excel in their ability to maneuver through intricate spaces, handle delicate operations, and interact with humans without posing risks. Continuum robots, with their elastic frameworks, exemplify this flexibility, capable of bending in virtually limitless ways thanks to their pliable backbones \cite{el2020nonlinear}. These robots are fabricated from diverse soft materials, including rubber \cite{ref1}, polyamide \cite{ref2}, silicone \cite{ref3}, and nitinol alloy (NiTi)\cite{ref4}, among others, to achieve their remarkable adaptability.

The inherent flexibility of soft robots plays a pivotal role in applications such as minimally invasive surgery \cite{ref5, ref6}, search and rescue missions \cite{samm2020developing, ahmed2022space}, and human-robot interaction \cite{seleem2023imitation}, where adapting to and navigating unpredictable environments is essential. Yet, the same compliant nature that offers these advantages also poses distinct challenges in terms of sensing and control \cite{shi2016shape}. With the increasing demand for soft robotics, there is a critical need to devise sensing systems that not only complement their flexible attributes but also provide the accurate feedback required for sophisticated tasks \cite{sorriento2019optical, bayoumy2014methods}.

A range of methods has been investigated for shape sensing in continuum robotic systems, encompassing fiber optic sensor-based approaches such as Fiber Bragg Grating (FBG) sensors \cite{ref14}, electrical impedance techniques \cite{ref15}, electromagnetic (EM) tracking \cite{ref16}, and imaging systems \cite{ref17}. Each of these techniques, however, comes with drawbacks that limit their effectiveness, especially in the context of Minimally Invasive Surgery (MIS), where the unique demands of the application challenge their suitability for integration with continuum robots.

While FBG sensors demonstrate notable shape sensing precision, their widespread adoption is curtailed by the high costs of both the fiber material and the interrogation equipment, posing a barrier to their practical use \cite{da2020challenges}. Similarly, electromagnetic positioning systems are commonly employed but encounter substantial challenges in surgical settings, where metal objects and various equipment can cause interference, compromising their accuracy \cite{sorriento2019optical}. Additionally, imaging methods like computer tomography and magnetic resonance imaging, despite their effectiveness, carry the disadvantage of exposing patients to ionizing radiation or X-rays, raising concerns over patient safety \cite{lubell2005drawbacks}.

Despite various existing methods for shape sensing, accurately determining the shape of continuum robots is still a difficult task. These methods need to be affordable and safe for humans while still being very accurate. Since most continuum robots do not use feedback for control, adding such feedback systems is complex. Thus, there is a significant need for a new method that is both flexible and cost-effective. To overcome the drawbacks of current stiff sensors, we propose a novel solution: the use of soft e-textile sensors to improve soft continuum robots' performance. Soft e-textile sensors are a new type of electronic fabric capable of detecting different physical properties around them and in the human body \cite{meena2023electronic}. Recently, these sensors have become increasingly popular for use in wearable devices and health monitoring \cite{du2022electronic}. This approach is not only adaptable and cost-saving but also overcomes the limitations of existing stiff sensors. By using the special features of soft e-textile sensors, we aim to greatly enhance the abilities of soft continuum robots, addressing an important gap in the field of robotics.

This research explores the possibility of employing e-textile sensors to broaden the range of force measurements the sensor can accurately detect. This enhancement is made possible through the sensor's surface conductivity characteristics. By incorporating the e-textile sensor directly into the robot's framework, we utilize its natural flexibility to monitor changes in resistance caused by the robot’s bending actions. To improve the accuracy of shape detection, we have devised a Convolutional Neural Network (CNN) approach driven by data. This CNN method processes the intricate patterns of resistance variations recorded by the sensors, enabling it to precisely determine the robot's shape and curvature in real-time. This advancement increases both the accuracy and adaptability of the robot in changing environments.

The structure of this paper is as follows. Section \ref{sec:design} describes the design of the e-textile sensor with stacked layers. Section \ref{sec:cnn} explains the deep CNN learning model used in sensing the robot's shape and facilitating shape control. Section \ref{sec:results} presents the experimental results demonstrating the use of the proposed soft sensor in sensing the shape of continuum robots. Finally, section \ref{sec:conclude} concludes this paper and suggests future research directions.

\section{E-textile Sensor Design}
\label{sec:design}

Our force sensor is crafted using EeonTex NW170-SPLPA, an elastic fabric produced by Eeonyx. This non-woven microfiber comprises 72\% nylon and 23\% spandex fibers and has been layered with a conductive polymer to create a piezoresistive effect. This implies that its electrical resistance changes when the material is contorted, allowing it to measure pressure, bending, stretching, or torsion precisely. The fabric's malleability, pliability, and capacity to deform under pressure make it an ideal option for shape sensor design.
\begin{figure}[!p!b]
	\centering
	\includegraphics[width=\linewidth]{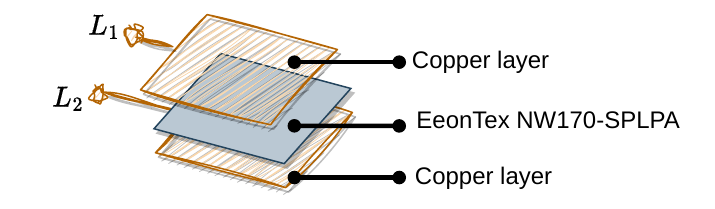}
	\caption{The sensor design, including the piezoresistive textile material and copper electrodes, is used to detect changes in pressure and strain.
		\label{fig:sensor_design} }
\end{figure}

To design the e-textile force sensor, we cut the textile material into a square with dimensions of $2$ cm by $2$ cm. The sensor layer is assembled between two copper electrodes arranged vertically with a piezoresistive pressure sensor, as depicted in Figure \ref{fig:sensor_design}. We then connected the electrodes to a breadboard and a resistance measuring circuit that employed a Wheatstone bridge signal conditioning and an Arduino UNO equipped with a 10-bit Analog-to-Digital Converter (ADC). The pressure applied to the material caused a change in its resistance.

\begin{figure}[!p!t]

	\centering
	\includegraphics[width=\linewidth]{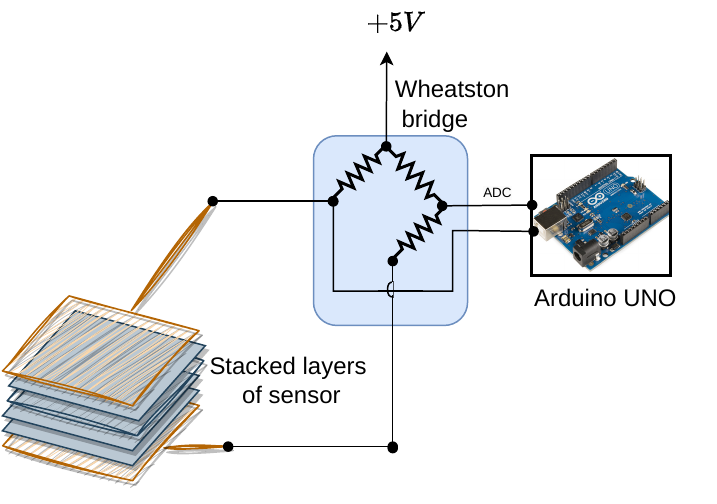}
	\caption{Illustration of the design of our soft force sensor with stacked layers.	\label{fig:stacked}}
\end{figure}

\begin{figure}[!p!t]
	\centering
	\includegraphics[width=.9\linewidth]{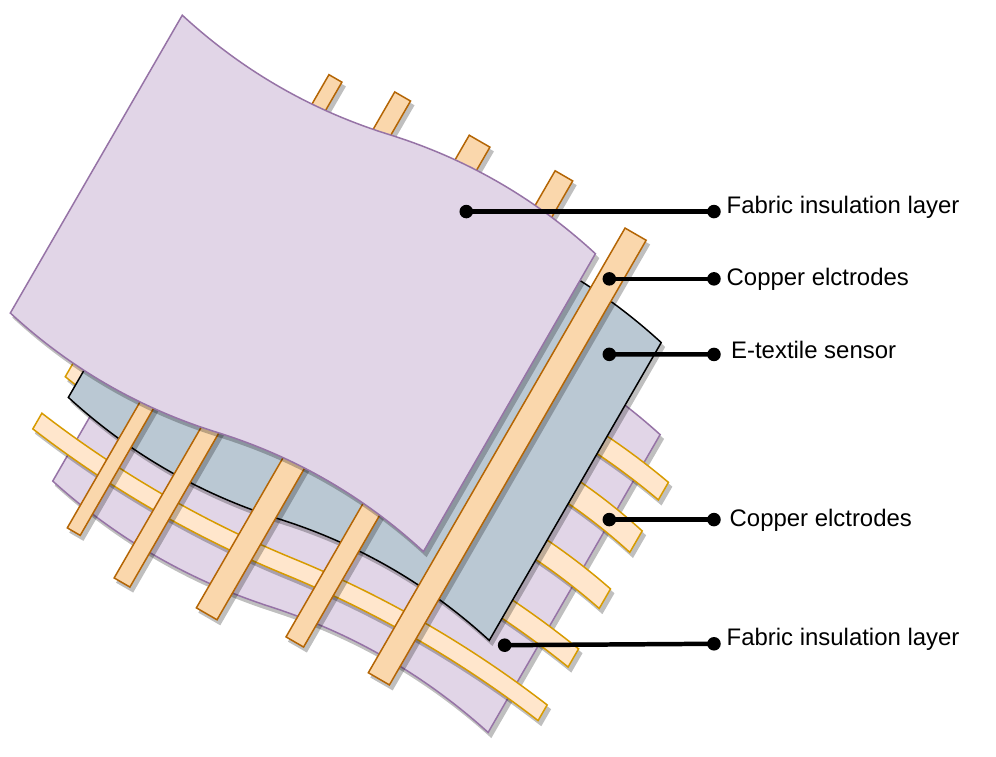}
	\caption{Illustration of the Soft E-Textile Sensor Interface: This diagram showcases the e-textile sensor at the core, flanked by copper electrodes, and encased within fabric insulation layers.	\label{fig:matrix}}

\end{figure}

To enhance the capacity of our soft sensor to measure a broader range of forces, a multi-layered approach is utilized, wherein each layer comprises a 2 cm by 2 cm square of conductive textile material. The first and last layers are connected to the copper electrodes, while the intermediate layers are sandwiched as illustrated in Figure \ref{fig:stacked}. Adding extra layers expands the sensor's dynamic range \cite{zhou2021textile, galeta2023design}, which may affect its response time in detecting applied forces. Achieving an optimal balance of layers based on response time and dynamic range requirements is critical for superior performance in particular applications. Such a balance presents a common challenge in the design of soft force sensors \cite{pyo2019multi}. To facilitate more sensing points on the e-textile shape sensing, a 4 $\times$ 4 conductors are arranged into a matrix above and below the soft sensing material, creating a set of 16 sensing points as depicted in Figure \ref{fig:matrix}.

\section{Deep-CNN Shape Sensing}\label{sec:cnn}
\begin{figure}[!p!t]

	\centering
	\includegraphics[width=.7\linewidth]{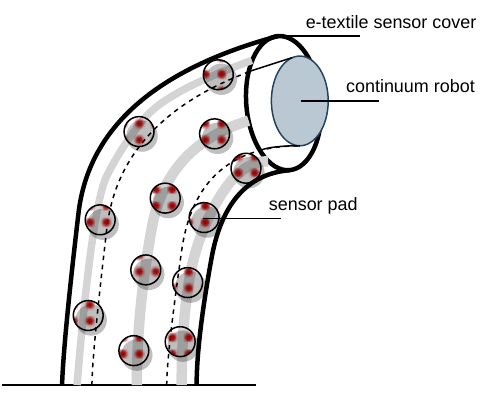}
	\caption{A one-section continuum robot is covered with the soft e-textile sensor comprising 16 sensing points arranged in 4 $\times$ 4 matrix. 	\label{fig:cover}}
\end{figure}

In the realm of soft robotics, the accurate prediction of robot shape is a critical aspect for effective functioning. In this research we employed a CNN deep learning model to approximate the curvature parameters of a soft robot based on the readings of the 16 sensing points of the e-textile soft sensor covering a one section continuum robot as highlighted in Figure \ref{fig:cover}. The choice of a CNN in this context is grounded in the model's inherent prowess in processing spatially correlated data representing the deformation of the soft robot. CNNs have demonstrated exceptional performance in various applications where spatial relationships are key, such as image and video recognition, medical image analysis, and natural language processing.

\subsection{Data Collection  and Processing}
Nonlinear regression models can be enhanced by increasing the size of the training dataset, resulting in more accurate predictions. However, this comes at the expense of substantial computational resources. Our research relies on dataset generated from experimental study of the robot with the e-textile sensor enveloped a cable-driven continuum robot. We applied a set of tension to one servo motor of the three motors actuating the robot and measured the e-textile 16 sensing points. To add the curvature $\kappa$ and the angle of curvature plan $\phi$ as the targets in our model, the robot's corresponding curvature is estimated from the measurement model with an Inertial Measurement Unit (IMU) sensor attached to the robot tip \cite{el2020nonlinear}.

It can be quite time-consuming to explore all the possible combinations of moments for the robot's tip. As a pragmatic approach, we generated a comprehensive dataset with 1330 data points. The collected data points were obtained within the expected workspace of the continuum robot, which has a length of 18 cm. The sampling process was meticulously designed to ensure thorough coverage of the operational range of the robot, while also adhering to permissible servo limits. In order to ensure reliable model performance and generalization, the dataset was divided into three subsets: 70\% for training, 15\% for validation, and the remaining 15\% for rigorous testing. This partitioning strategy was implemented to aid in achieving robust model training and evaluation.

In order to utilize CNNs for shape sensing of soft robots, we must first convert each data sample, represented as the sensor readings $p_i \in \mathds{R}^{16}$, into a 2D feature image, are denoted as $\mathbf{I} \in \mathds{R}^{1330 \times 4 \times 4}$. This transformation is not merely a reshaping of data but a critical step in enabling the CNN to interpret the spatial correlations between different sensing points on the robot's surface. 

To enhance the model's training stability, the training and testing data are normalized using the Z-score technique, which ensures that our data has a mean of $\mu = 0$ and a standard deviation of $\sigma = 1$. Each data point $\mathbf{p}$ is normalized as follows:

$$ \mathbf{p}\prime= (\mathbf{p} - \mu)/\sigma $$

\noindent It is important to note that the values of $\mu$ and $\sigma$ are calculated solely based on the training data.

\subsection{CNN Model Architectures}
\begin{figure*}[!t!p]%
	\centering
	\label{fig:n1}\includegraphics[width=\textwidth]{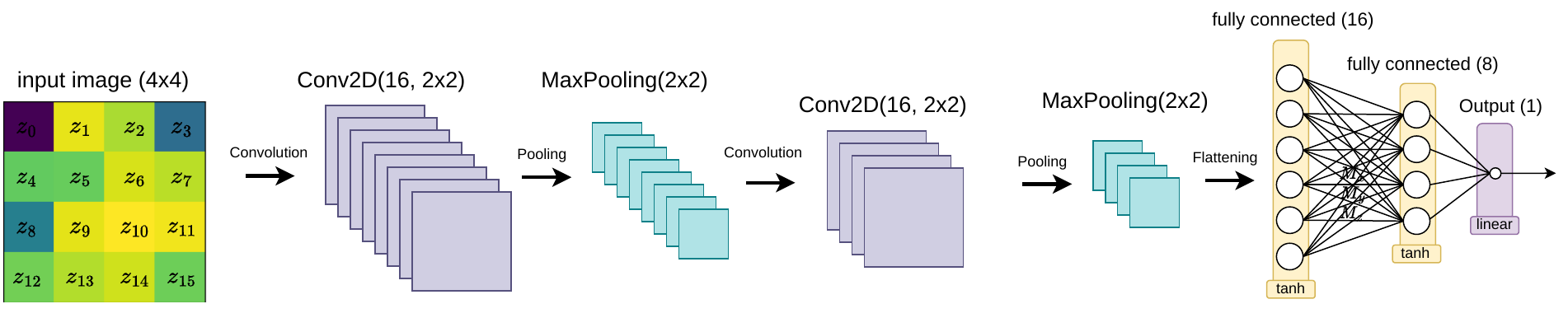}
	\caption{The CNN architectures employed in estimating the curvature of the soft robot from the sensor readings. \label{fig:network}}
\end{figure*}

The CNN architecture employed in our study is designed to perform regression tasks, given that our objective is to quantify the curvature of the soft robot rather than classify it into discrete categories. This regression output is a continuous value representing the curvature $\kappa$ and angle of curvature plan $\phi$ that are estimated form an IMU attached to the robot's tip during data collection. The network achieves this through two convolutional layers, each responsible for extracting increasingly abstract features from the input image. The kernel size of first convolution layer is chosen to 16 and 8 while the hyperbolic tangent (tanh) activation function is selected. These layers are followed by two fully connected layers of 16 and 8 units that integrate these features to predict the robot's curvature as shown in Figure \ref{fig:network}.

\subsection{Training of CNN model}
In the training process of our CNN models, the Mean-Squared Error (MSE) loss function has been utilized, playing a pivotal role in the deep regressor model that approximates the shape parameters of the continuum robot. The MSE loss function is instrumental in quantifying the discrepancy between the CNN model's predicted output, represented as $\bm{\hat{y}}(\zeta)$, and the actual reference output $\bm{y}$, which corresponds to the robot's curvature. This choice of loss function is aligned with the regression nature of our task, where the precise quantification of errors in continuous output values is essential. The Mean-Squared Error is mathematically formulated as:

\begin{equation}
	\label{eq:mse}
	\text{MSE} = \dfrac{1}{N}  \sum_{i=0}^{N-1} [\bm{\hat{y}}(\zeta)^i -\bm{y}^i ]^2
\end{equation}

\noindent The symbol $\zeta$ in the given equation denotes the network parameters, encompassing learnable weights and biases. To enhance the model's performance, the Adam optimizer \cite{kingma2014adam} has been employed, with an initial learning rate of $10^{-2}$ as its hyper-parameter. Furthermore, customization of hyper-parameters is warranted for individual numerical examples, such as the training batch size of 32 and 500 epochs.
\section{Results and Discussion}\label{sec:results}
\begin{figure}[!p!b]
	\centering
	\includegraphics[width=\linewidth, trim=0cm 0cm 3.5cm 0cm, clip]{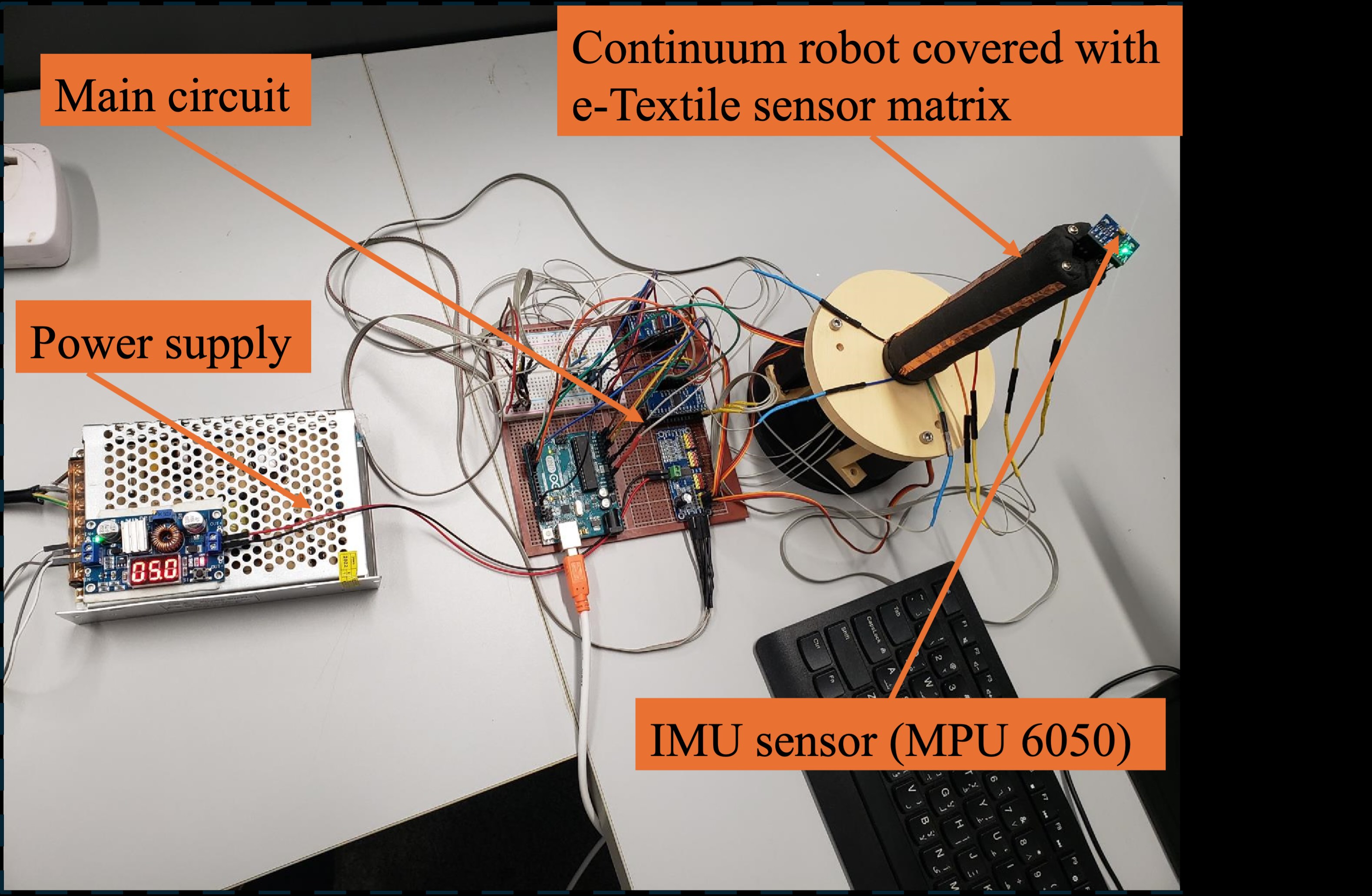}
	\caption{The experimental setup of a cable-driven continuum robot, IMU, soft e-textile sensor, and Arduino UNO for estimating the shape of the continuum robot.} \label{fig:exp}
\end{figure}
In this section, we explore the performance of the proposed CNN-based approach for shape sensing in continuum robots, leveraging a soft e-textile sensor. Our investigation begins with an evaluation of the e-textile sensor array's capacity to dynamically capture the robot's shape changes through a 4x4 matrix of sensing points highlighted in Figure \ref{fig:exp}. Each point's electrical resistance is inversely related to the pressure applied by the robot's deformation, offering a nuanced understanding of its structural changes.

Following the sensor evaluation, our attention shifts to the performance of the CNN model. By closely monitoring the training and validation losses, we identify critical areas for enhancement, ensuring the model's robustness and effectiveness. The examination extends to testing the model against a distinct set of data, affirming its reliability and precision in predicting the robot's curvature and the angle of curvature plan.

Moreover, to ascertain the most effective model architecture, we employed a 5-fold cross-validation technique on five different models. This rigorous analysis facilitated a deep understanding of each model's strengths and weaknesses, guiding us towards the most suitable architecture for our application.

Through these methodical experiments, we aim to refine the integration of CNN-based shape sensing with soft e-textile sensors, pushing the boundaries of what's achievable in the realm of continuum robotics.

\subsection{e-Textile Sensor Responsiveness}
\begin{figure}[!p!t]
	\centering
	\includegraphics[width=.6\linewidth]{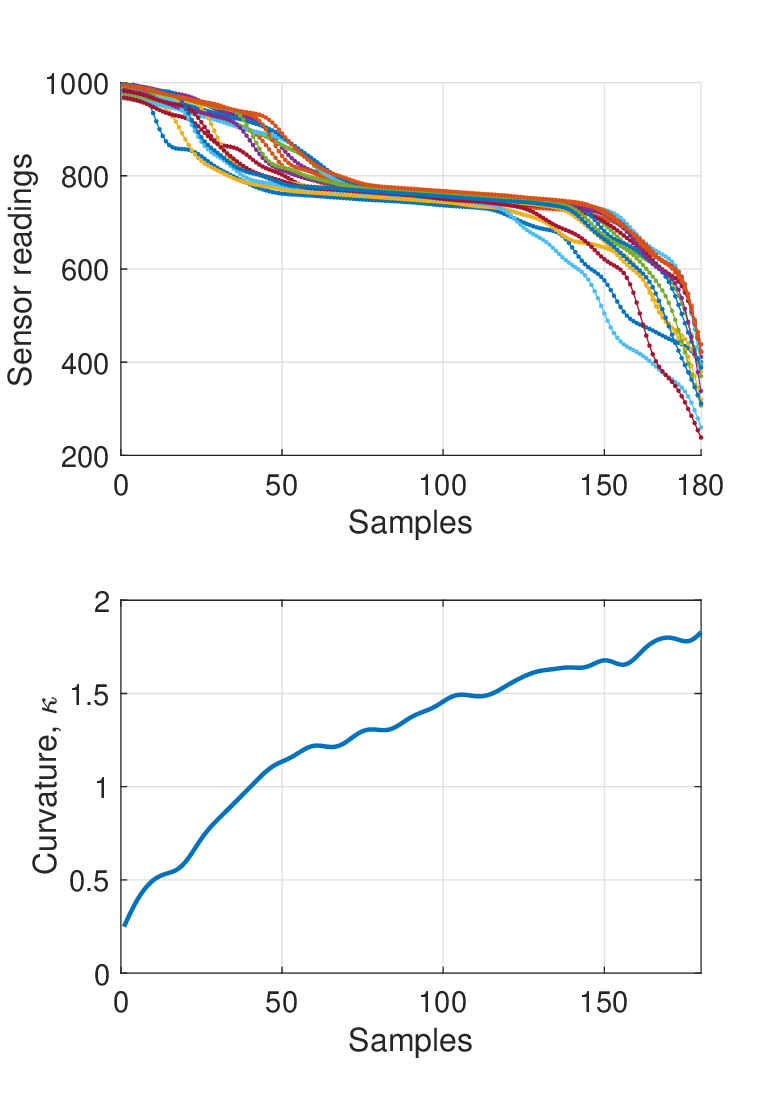}
	\caption{The change of the sensors' readings of a 4x4 matrix of e-textile sensing points capturing the continuum robot's deformation with respect to change in the robot's curvature $\kappa$.} \label{fig:modes}
\end{figure}

This experiment investigates the efficacy of an e-textile sensor array in dynamically sensing the shape of a continuum robot through a 4x4 matrix of sensing points, where each point's resistance inversely correlates with the applied pressure resulting from the robot's deformation in terms of curvature $\kappa$. As depicted in Figure \ref{fig:modes}, our findings illuminate the sensor's sensitivity and precision in mapping curvature changes, providing essential insights into the integration of soft robotics and the soft e-textile sensor.

The data reveal that resistance changes at the sensing points are not uniform but vary in a manner that directly reflects the robot's deformation pattern. This variation suggests that the e-textile sensor can not only detect the presence of bending but also discern the bending's magnitude and directionality.

\subsection{CNN Model Performance}
\begin{figure}[!t!p]
	\centering
		\includegraphics[width=.9\columnwidth]{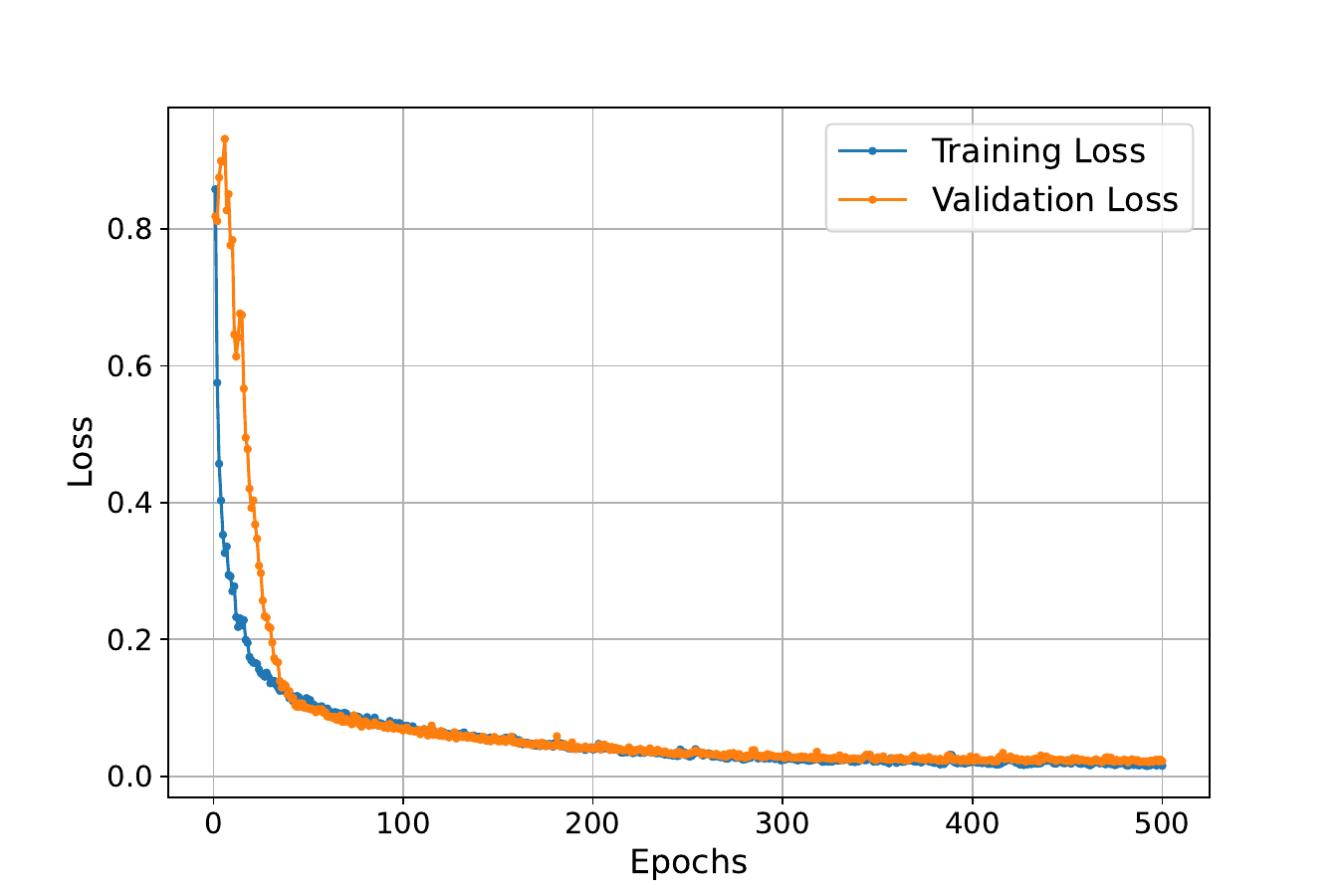}
	\caption{Graph showing the Mean Squared Error (MSE) of training and validation losses during the training of the CNN-based shape sensing of continuum robots over 500 epochs.\label{fig:losses}}
\end{figure}

The training and prediction tasks were performed using the TensorFlow 2.x API on a desktop computer. The selection of computer hardware notably influences the training time. In this research, a CPU with a 2.8 GHz clock speed facilitated the model training process. Figure \ref{fig:losses} presents the Mean Squared Error (MSE) losses for the CNN model over 500 epochs, illustrating their progressive convergence to minimal loss values within both the training and validation datasets.

\begin{figure}[!p!t]%
	\centering
	\includegraphics[width=.8\columnwidth]{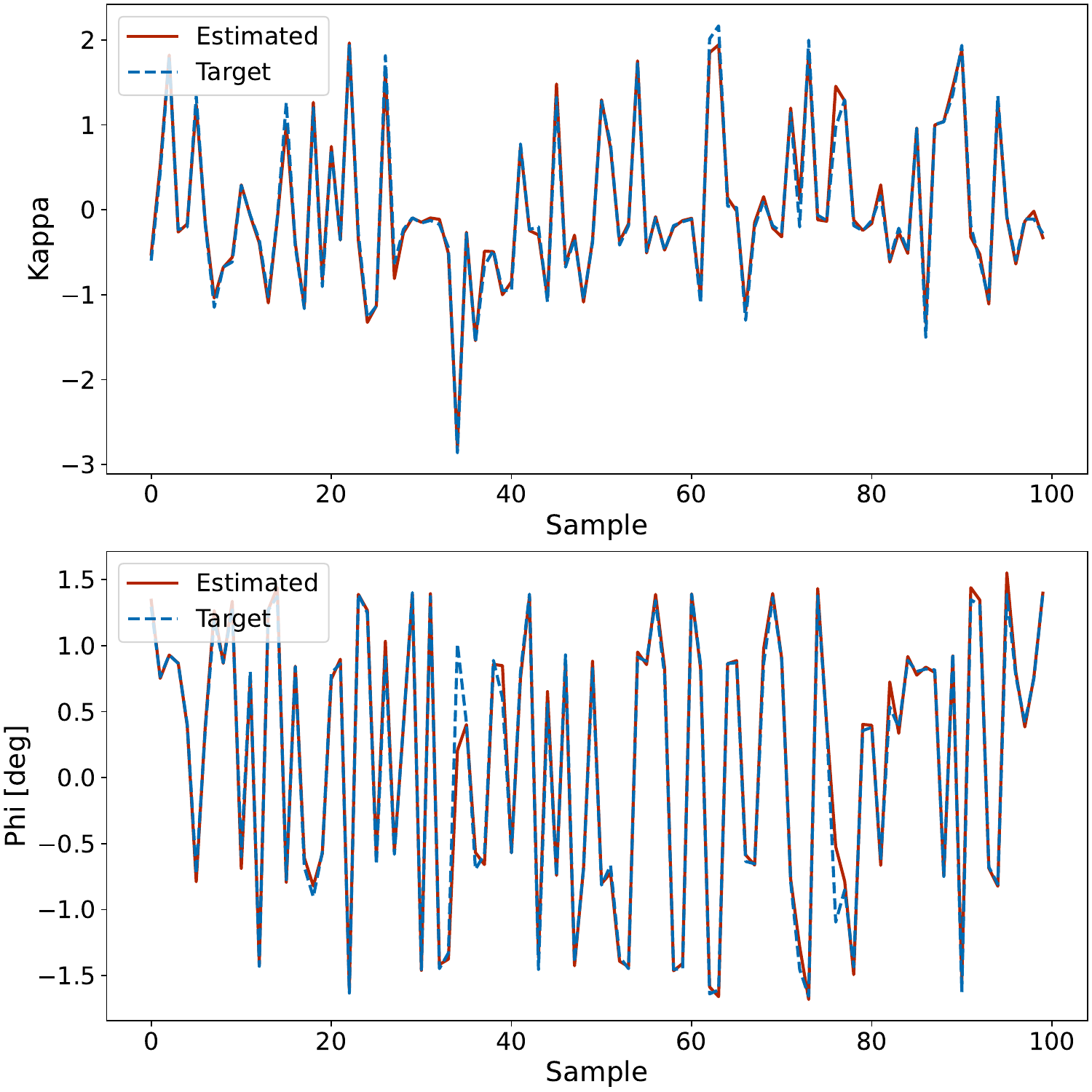}
	\caption{A comparison of target versus estimated validation dataset of the robot's parameters $\kappa$ and $\phi$ of the CNN-based shape sensing model,  }\label{fig:error}
\end{figure}

To assess the models' efficacy, we utilized a testing dataset of 400 samples, processed in batches of 64. Figure \ref{fig:error} contrasts the CNN-based model's predictions with the actual target values, highlighting the absolute error. Owing to space limitations, only the initial 100 samples are shown. The negligible error margin of the CNN model is demonstrated by the tight alignment of the estimated outputs with the target signals.

\subsection{K-Fold Cross Validation}
In this subsection, we conducted a thorough evaluation of the proposed CNN model using the K-Fold Cross-Validation technique, as described in \cite{anguita2009k}. This approach is crucial for thoroughly examining the model's ability to sense shape across various architectural designs, ensuring its effectiveness and reliability. The dataset was divided into $k=5$ separate subsets to support a cyclic process of model training and evaluation. This detailed evaluation method provides insight into the model's performance across different conditions. Importantly, this framework not only assists in identifying the most effective architectural design but also sheds light on the model's ability to generalize in real-world applications.

\begin{figure}[!t!p]
	\centering
	\begin{minipage}{\columnwidth}
		\centering
				\subfloat[]{\label{fig:folds1}\includegraphics[width=.65\textwidth]{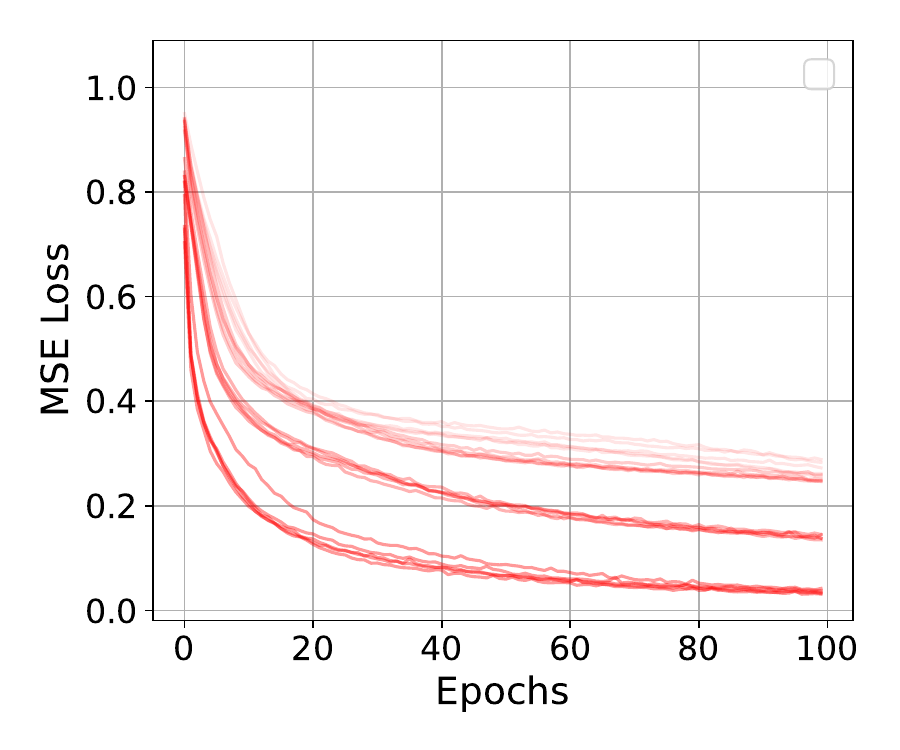}}
	\end{minipage}%
	
	\begin{minipage}{\columnwidth}
		\centering
			\subfloat[]{\label{fig:folds2}\includegraphics[width=.65\textwidth]{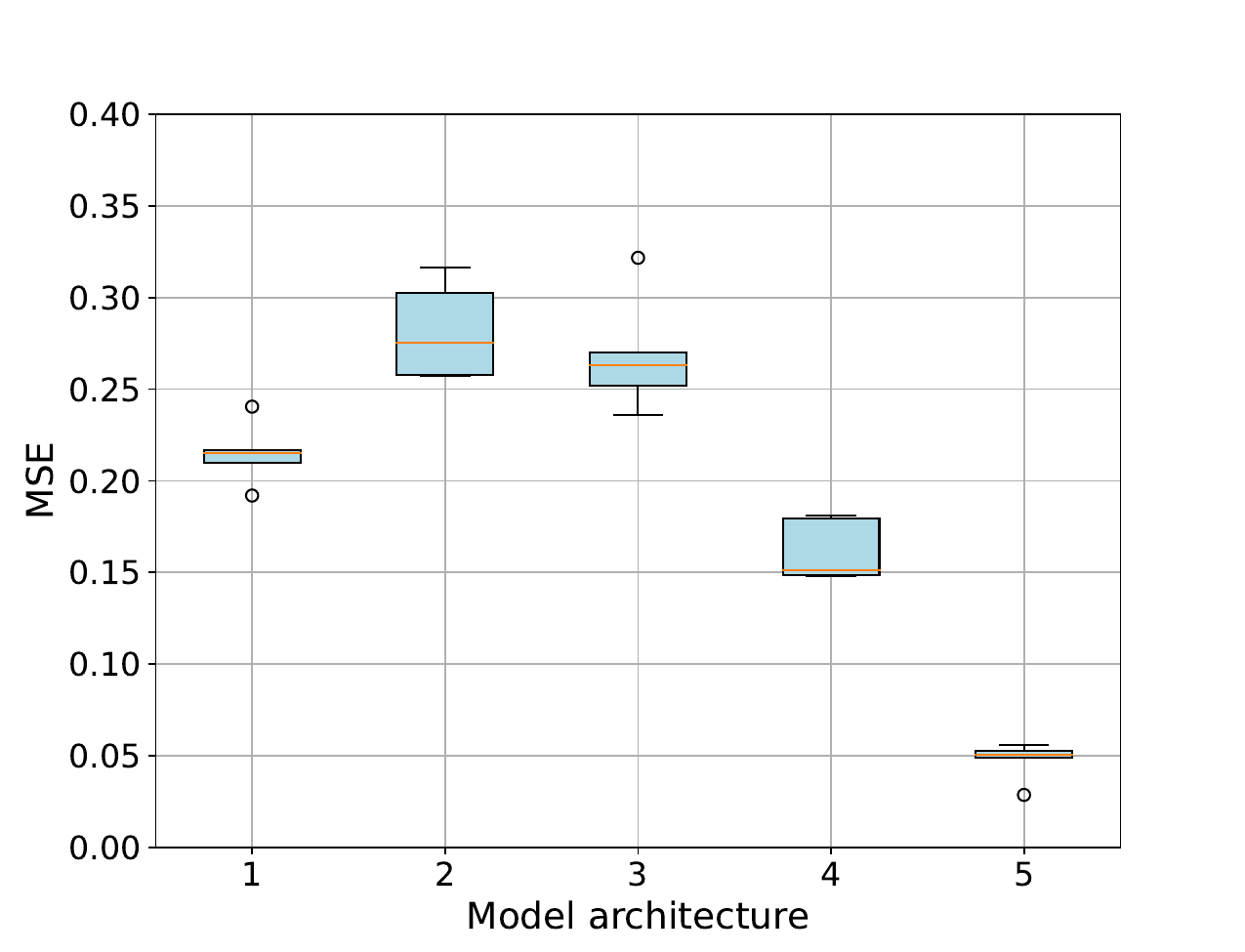}}
	\end{minipage}%
	
	\caption{(a) Training losses expressed as mean squared error (MSE) for five deep neural network (DNN) models during 5-fold cross-validation with (b) mean and standard deviations. \label{fig:kfold}}
\end{figure}

In this study, we developed a suite of five Deep Neural Network (DNN) models, incorporating both feed-forward and convolutional neural network (CNN) architectures. These models span a spectrum of trainable parameters, ranging from 259 to 2515, and are organized into 2 to 3 layers, detailed in Table \ref{tab:arch}. The convolutional layers are designated by a 'C' prefix, followed by a subscript that specifies the number of kernels and their size. Fully connected layers are marked with an 'F' prefix and a subscript denoting the neuron count. A uniform stride of 1 was applied across convolutional layers, with padding excluded. The activation functions were mainly the hyperbolic tangent (tanh), with the exception of the output layer, which employed a linear activation function. If a model did not incorporate a convolutional layer at the beginning, a flatten layer was introduced to convert the input pose image into a column vector, ensuring compatibility with the network's input requirements.
\begin{table}[!p!t]
	\centering
	\caption{Selected architectures for the 5-fold cross validation experiment. \label{tab:arch}}
	\begin{tabular}{|c|l|c|c|}
		\hline
		\textbf{Model} & \textbf{Layers} & \textbf{Size} & \textbf{Average MSE}\\
		\hline
		1 & $F_{16}, F_{3}$ & 306 & 0.21\\
		2 & $C_{8,2}, F_{3}$  & 186 & 0.28\\
		3 & $C_{8,2}, C_{4,2}, F_{3}$    & 206 & 0.26\\
		4 & $C_{16,2}, C_{8,2}, F_{3}$   & 666 & 0.16\\
		5 & $C_{32,2}, C_{16,2}, F_{8}, F_{3}$  & 2762 & 0.04\\
		\hline
	\end{tabular}
\end{table}
These models underwent training over 100 epochs, utilizing the Adam optimizer with Mean Squared Error (MSE) as the primary loss metric. To enhance the thoroughness of evaluation and validation, a 5-fold cross-validation method was employed. The learning algorithm's convergence trends, covering all network configurations and validation folds, are illustrated in Figure \ref{fig:kfold}-a. Moreover, Figure \ref{fig:kfold}-b offers an in-depth examination of the average MSE losses and their standard deviations across different model architectures, with a particular emphasis on the CNN models.

Significantly, networks featuring a higher parameter count are highlighted in dark red, signifying that more extensive networks tend to converge more effectively towards lower MSE values. For the sake of simplification in our selection methodology, we chose a neural network that demonstrated commendable performance throughout both training and testing stages. Investigating architectures that are both compact and capable of rapid learning offers potential to enhance the robustness of the overall system.

\section{Conclusion}\label{sec:conclude}
In summary, this study presents a novel soft e-textile resistive sensor, utilizing a deep convolutional neural network (CNN) specifically engineered for the real-time shape sensing of soft continuum robots. The CNN model is meticulously crafted to navigate the spatial continuities and intrinsic nonlinear characteristics of sensor data, thereby enhancing the capability to accurately discern complex shapes in soft continuum robotics. It exhibits exceptional precision in determining the robot's curvature and the planar angle of curvature.

However, further research is required to extend this method to the complex shape sensing, involving multiple sections of continuum robots. Meanwhile, investigate the optimization of e-textile sensor configurations to improve sensitivity, durability, and response time could be one of the future works to be added. This includes exploring different materials, fabrication methods, and sensor geometries.


\bibliographystyle{ieeetr}
\bibliography{references.bib}

\begin{thebibliography}{10}

\bibitem{el2020nonlinear}
H.~El-Hussieny, I.~A. Hameed, and J.-H. Ryu, ``Nonlinear model predictive
  growth control of a class of plant-inspired soft growing robots,'' {\em IEEE
  Access}, vol.~8, pp.~214495--214503, 2020.

\bibitem{seleem2023recent}
I.~A. Seleem, H.~El-Hussieny, and H.~Ishii, ``Recent developments of actuation
  mechanisms for continuum robots: a review,'' {\em International Journal of
  Control, Automation and Systems}, vol.~21, no.~5, pp.~1592--1609, 2023.

\bibitem{ref1}
A.~D. Marchese, R.~K. Katzschmann, and D.~Rus, ``A recipe for soft fluidic
  elastomer robots,'' {\em Soft robotics}, vol.~2, no.~1, pp.~7--25, 2015.

\bibitem{ref2}
A.~Bajo and N.~Simaan, ``Hybrid motion/force control of multi-backbone
  continuum robots,'' {\em The International journal of robotics research},
  vol.~35, no.~4, pp.~422--434, 2016.

\bibitem{ref3}
T.~Zheng, D.~T. Branson, E.~Guglielmino, R.~Kang, G.~A. Medrano~Cerda,
  M.~Cianchetti, M.~Follador, I.~S. Godage, and D.~G. Caldwell, ``Model
  validation of an octopus inspired continuum robotic arm for use in underwater
  environments,'' {\em Journal of Mechanisms and Robotics}, vol.~5, no.~2,
  p.~021004, 2013.

\bibitem{ref4}
L.~Wang and N.~Simaan, ``Geometric calibration of continuum robots: Joint space
  and equilibrium shape deviations,'' {\em IEEE Transactions on Robotics},
  vol.~35, no.~2, pp.~387--402, 2019.

\bibitem{ref5}
T.~Kato, I.~Okumura, S.-E. Song, A.~J. Golby, and N.~Hata, ``Tendon-driven
  continuum robot for endoscopic surgery: Preclinical development and
  validation of a tension propagation model,'' {\em IEEE/ASME Transactions on
  Mechatronics}, vol.~20, no.~5, pp.~2252--2263, 2014.

\bibitem{ref6}
X.~Zhang, W.~Li, P.~W.~Y. Chiu, and Z.~Li, ``A novel flexible robotic endoscope
  with constrained tendon-driven continuum mechanism,'' {\em IEEE Robotics and
  Automation Letters}, vol.~5, no.~2, pp.~1366--1372, 2020.

\bibitem{samm2020developing}
R.~T. Samm, V.~Durairajah, and S.~Gobee, ``Developing a fully soft robotic
  snake for search and rescue,'' {\em Solid State Technology}, vol.~63, no.~1s,
  pp.~1314--1329, 2020.

\bibitem{ahmed2022space}
A.~Ahmed, A.~Maged, A.~Soliman, H.~El-Hussieny, and M.~Magdy, ``Space
  deformation based path planning for mobile robots,'' {\em ISA transactions},
  vol.~126, pp.~666--678, 2022.

\bibitem{seleem2023imitation}
I.~A. Seleem, H.~El-Hussieny, and H.~Ishii, ``Imitation-based motion planning
  and control of a multi-section continuum robot interacting with the
  environment,'' {\em IEEE Robotics and Automation Letters}, vol.~8, no.~3,
  pp.~1351--1358, 2023.

\bibitem{shi2016shape}
C.~Shi, X.~Luo, P.~Qi, T.~Li, S.~Song, Z.~Najdovski, T.~Fukuda, and H.~Ren,
  ``Shape sensing techniques for continuum robots in minimally invasive
  surgery: A survey,'' {\em IEEE Transactions on Biomedical Engineering},
  vol.~64, no.~8, pp.~1665--1678, 2016.

\bibitem{sorriento2019optical}
A.~Sorriento, M.~B. Porfido, S.~Mazzoleni, G.~Calvosa, M.~Tenucci, G.~Ciuti,
  and P.~Dario, ``Optical and electromagnetic tracking systems for biomedical
  applications: A critical review on potentialities and limitations,'' {\em
  IEEE reviews in biomedical engineering}, vol.~13, pp.~212--232, 2019.

\bibitem{bayoumy2014methods}
A.~Bayoumy, A.~Nada, and S.~Megahed, ``Methods of modeling slope
  discontinuities in large size wind turbine blades using absolute nodal
  coordinate formulation,'' {\em Proceedings of the Institution of Mechanical
  Engineers, Part K: Journal of Multi-body Dynamics}, vol.~228, no.~3,
  pp.~314--329, 2014.

\bibitem{ref14}
I.~Floris, J.~M. Adam, P.~A. Calder{\'o}n, and S.~Sales, ``Fiber optic shape
  sensors: A comprehensive review,'' {\em Optics and Lasers in Engineering},
  vol.~139, p.~106508, 2021.

\bibitem{ref15}
J.~Avery, M.~Runciman, A.~Darzi, and G.~P. Mylonas, ``Shape sensing of variable
  stiffness soft robots using electrical impedance tomography,'' in {\em 2019
  International Conference on Robotics and Automation (ICRA)}, pp.~9066--9072,
  IEEE, 2019.

\bibitem{ref16}
S.~Song, Z.~Li, H.~Yu, and H.~Ren, ``Electromagnetic positioning for tip
  tracking and shape sensing of flexible robots,'' {\em IEEE Sensors Journal},
  vol.~15, no.~8, pp.~4565--4575, 2015.

\bibitem{ref17}
M.~Wagner, S.~Schafer, C.~Strother, and C.~Mistretta, ``4d interventional
  device reconstruction from biplane fluoroscopy,'' {\em Medical physics},
  vol.~43, no.~3, pp.~1324--1334, 2016.

\bibitem{da2020challenges}
T.~da~Veiga, J.~H. Chandler, P.~Lloyd, G.~Pittiglio, N.~J. Wilkinson, A.~K.
  Hoshiar, R.~A. Harris, and P.~Valdastri, ``Challenges of continuum robots in
  clinical context: a review,'' {\em Progress in Biomedical Engineering},
  vol.~2, no.~3, p.~032003, 2020.

\bibitem{lubell2005drawbacks}
D.~L. Lubell, ``Drawbacks and limitations of computed tomography,'' {\em Texas
  Heart Institute Journal}, vol.~32, no.~2, p.~250, 2005.

\bibitem{meena2023electronic}
J.~S. Meena, S.~B. Choi, S.-B. Jung, and J.-W. Kim, ``Electronic textiles: New
  age of wearable technology for healthcare and fitness solutions,'' {\em
  Materials Today Bio}, p.~100565, 2023.

\bibitem{du2022electronic}
K.~Du, R.~Lin, L.~Yin, J.~S. Ho, J.~Wang, and C.~T. Lim, ``Electronic textiles
  for energy, sensing, and communication,'' {\em IScience}, vol.~25, no.~5,
  2022.

\bibitem{zhou2021textile}
Z.~Zhou, N.~Chen, H.~Zhong, W.~Zhang, Y.~Zhang, X.~Yin, and B.~He,
  ``Textile-based mechanical sensors: A review,'' {\em Materials}, vol.~14,
  no.~20, p.~6073, 2021.

\bibitem{galeta2023design}
E.~V. Galeta, S.~Ahmed, V.~Parque, and H.~El-Hussieny, ``Design and
  characterization of an e-textile sensor for shape sensing of soft continuum
  robots,'' in {\em 2023 62nd Annual Conference of the Society of Instrument
  and Control Engineers (SICE)}, pp.~1110--1115, IEEE, 2023.

\bibitem{pyo2019multi}
S.~Pyo, J.~Lee, W.~Kim, E.~Jo, and J.~Kim, ``Multi-layered, hierarchical
  fabric-based tactile sensors with high sensitivity and linearity in ultrawide
  pressure range,'' {\em Advanced Functional Materials}, vol.~29, no.~35,
  p.~1902484, 2019.

\bibitem{kingma2014adam}
D.~P. Kingma and J.~Ba, ``Adam: A method for stochastic optimization,'' {\em
  arXiv preprint arXiv:1412.6980}, 2014.

\bibitem{anguita2009k}
D.~Anguita, A.~Ghio, S.~Ridella, and D.~Sterpi, ``K-fold cross validation for
  error rate estimate in support vector machines.,'' in {\em DMIN},
  pp.~291--297, 2009.

\end{thebibliography}

\end{document}